\documentclass[10pt,twocolumn,letterpaper]{article} 
\usepackage[svgnames,table]{xcolor}
\usepackage{cvpr}
\usepackage{times}
\usepackage{epsfig}
\usepackage{graphicx}
\usepackage{color}
\usepackage[pagebackref=true,breaklinks=true,letterpaper=true,colorlinks,bookmarks=false]{hyperref}
\usepackage{amsmath,amsfonts,amssymb,mathrsfs}
\usepackage{makecell}

\usepackage{graphicx}
\usepackage[ruled]{algorithm2e}
\usepackage{amsmath,amssymb}
\usepackage{epstopdf}
\usepackage{tabularx}
\usepackage{eqnarray}
\usepackage{array,booktabs}
\usepackage{enumitem}
\usepackage[space]{grffile} 
\usepackage{soul} 
\usepackage{float}
\usepackage{tabularx,ragged2e} 
\usepackage{bbm}
\usepackage{pifont} 
\usepackage{cancel}
\usepackage{wrapfig}
\usepackage{color, colortbl}

\makeatletter

\renewcommand{\paragraph}{%
  \@startsection{paragraph}{4}%
  {\z@}{0.3ex \@plus 1ex \@minus .2ex}{-1em}
  {\normalfont\normalsize\bfseries}%
}
\makeatother

\definecolor{myGray}{rgb}{0.6,0.6,0.6}
\definecolor{myBlue}{rgb}{0.1,0.1,0.8}

\DeclareMathOperator*{\argmax}{arg\,max}

\def\onedot{\ifx\@let@token.\else.\null\fi\xspace}

\def\eg{\emph{e.g}\onedot} 
\def\ie{\emph{i.e}\onedot}

\def\etal{\emph{et al}\onedot}

\definecolor{pcGray}{rgb}{0.5,0.5,0.5}

\definecolor{pccGray}{rgb}{0.3,0.3,0.3}

\newcommand{\fig}{Fig.~}
\newcommand{\eq}{Eq.\,}
\newcommand{\sect}{Section~}
\newcommand{\tab}{Table~}

\newcommand{\hangBoxC}[1]{%
  \begin{minipage}[c]{\textwidth}\begin{tabbing} 
  ~\\[-\baselineskip] 
  #1 
  \end{tabbing}
  \end{minipage}}

\newcommand\minput[1]{%
  \input{#1}%
  \ifhmode\ifnum\lastnodetype=11 \unskip\fi\fi}

\usepackage{listings}
\usepackage{color}

\ifx \@etal \@empty \newcommand{\etal}{et al.} \fi
\ifx \@eg \@empty \newcommand{\eg}{e.g.,~} \fi
\ifx \@ie \@empty \newcommand{\ie}{i.e.,~} \fi
\ifx \@etc \@empty  \fi

\newcommand{\bd}{\boldsymbol{d}}

\newcommand{\bh}{{\boldsymbol{h}}}

\newcommand{\bq}{{\boldsymbol{q}}}

\newcommand{\bs}{{\boldsymbol{s}}}

\newcommand{\bv}{{\boldsymbol{v}}}

\newcommand{\bx}{{\boldsymbol{x}}}

\cvprfinalcopy
\def\cvprPaperID{340}

\ifcvprfinal\fi

\newcommand{\vvspace}[1]{\vspace{#1}}

\newcommand{\loss}{\mathcal{L}}
\newcommand{\setSu}{\mathcal{S}}
\newcommand{\setTr}{\mathcal{T}}
\newcommand{\setSuTr}{\mathcal{S}^\textrm{tr}}

\graphicspath{{visualizations/}}

\newcommand{\vis}[1]{
  \hangBoxC{\includegraphics[width=30mm]{#1_padded.jpg}\\\vspace{3pt}\\\begin{minipage}{30mm}\centering\footnotesize\input{#1.txt}~?\\Correct answer:~\input{#1_ans0.txt}.\end{minipage}}& 
  \vspace{1pt}\hangBoxC{\hspace{3.5mm}\includegraphics[width=27mm]{#1_1_supp-8_padded.jpg}\\\begin{minipage}{34mm}\centering\scriptsize\input{#1_1_supp-8.txt}.\\\vspace{4pt}\end{minipage}\\\hspace{3.5mm}\includegraphics[width=27mm]{#1_4_supp-8_padded.jpg}\\\begin{minipage}{34mm}\centering\scriptsize\input{#1_4_supp-8.txt}.\end{minipage}}& 
  \vspace{1pt}\hangBoxC{\hspace{3.5mm}\includegraphics[width=27mm]{#1_2_supp-8_padded.jpg}\\\begin{minipage}{34mm}\centering\scriptsize\input{#1_2_supp-8.txt}.\\\vspace{4pt}\end{minipage}\\\hspace{3.5mm}\includegraphics[width=27mm]{#1_5_supp-8_padded.jpg}\\\begin{minipage}{34mm}\centering\scriptsize\input{#1_5_supp-8.txt}.\end{minipage}}& 
  \vspace{1pt}\hangBoxC{\hspace{3.5mm}\includegraphics[width=27mm]{#1_3_supp-8_padded.jpg}\\\begin{minipage}{34mm}\centering\scriptsize\input{#1_3_supp-8.txt}.\\\vspace{4pt}\end{minipage}\\\hspace{3.5mm}\includegraphics[width=27mm]{#1_6_supp-8_padded.jpg}\\\begin{minipage}{34mm}\centering\scriptsize\input{#1_6_supp-8.txt}.\end{minipage}}& 
  \begin{minipage}{30mm}
    \footnotesize
    ~Without adaptation:
    \begin{itemize}[topsep=0.4ex,itemsep=-0.8ex,partopsep=0ex,parsep=0.3ex,label={},leftmargin=0.9ex]
      \input{#1_ans1.txt}
    \end{itemize}\vspace{5pt}
    ~After adaptation:
    \begin{itemize}[topsep=0.4ex,itemsep=-0.8ex,partopsep=0ex,parsep=0.3ex,label={},leftmargin=0.9ex]
      \input{#1_ans2.txt}
    \end{itemize}
  \end{minipage}\\
}

\newcommand{\visCapt}[1]{
  \hangBoxC{\includegraphics[width=30mm]{#1_padded.jpg}\\\vspace{3pt}\\\begin{minipage}{30mm}\centering\footnotesize\input{#1.txt}~?\\Correct answer:~\input{#1_ans0.txt}.\end{minipage}}& 
  \vspace{1pt}\hangBoxC{\hspace{3.5mm}\includegraphics[width=27mm]{#1_1_supp-1_padded.jpg}\\\begin{minipage}{34mm}\centering\scriptsize\input{#1_1_supp-1.txt}\\\vspace{4pt}\end{minipage}\\\hspace{3.5mm}\includegraphics[width=27mm]{#1_4_supp-1_padded.jpg}\\\begin{minipage}{34mm}\centering\scriptsize\input{#1_4_supp-1.txt}\end{minipage}}& 
  \vspace{1pt}\hangBoxC{\hspace{3.5mm}\includegraphics[width=27mm]{#1_2_supp-1_padded.jpg}\\\begin{minipage}{34mm}\centering\scriptsize\input{#1_2_supp-1.txt}\\\vspace{4pt}\end{minipage}\\\hspace{3.5mm}\includegraphics[width=27mm]{#1_5_supp-1_padded.jpg}\\\begin{minipage}{34mm}\centering\scriptsize\input{#1_5_supp-1.txt}\end{minipage}}& 
  \vspace{1pt}\hangBoxC{\hspace{3.5mm}\includegraphics[width=27mm]{#1_3_supp-1_padded.jpg}\\\begin{minipage}{34mm}\centering\scriptsize\input{#1_3_supp-1.txt}\\\vspace{4pt}\end{minipage}\\\hspace{3.5mm}\includegraphics[width=27mm]{#1_6_supp-1_padded.jpg}\\\begin{minipage}{34mm}\centering\scriptsize\input{#1_6_supp-1.txt}\end{minipage}}& 
  \begin{minipage}{30mm}
    \footnotesize
    ~Without adaptation:
    \begin{itemize}[topsep=0.4ex,itemsep=-0.8ex,partopsep=0ex,parsep=0.3ex,label={},leftmargin=0.9ex]
      \input{#1_ans1.txt}
    \end{itemize}\vspace{5pt}
    ~After adaptation:
    \begin{itemize}[topsep=0.4ex,itemsep=-0.8ex,partopsep=0ex,parsep=0.3ex,label={},leftmargin=0.9ex]
      \input{#1_ans2.txt}
    \end{itemize}
  \end{minipage}\\
}

\makeatletter
\let\@fnsymbol\@arabic
\makeatother

\begin{document}

\title{Actively Seeking and Learning from Live Data}

\author{Damien Teney ~~~~~ Anton van den Hengel\\
Australian Institute for Machine Learning\\
The University of Adelaide\\
Adelaide, Australia\\
{\tt\small \{damien.teney,anton.vandenhengel\}@adelaide.edu.au}
}

\maketitle
\thispagestyle{empty} 

\begin{abstract}
One of the key limitations of traditional machine learning methods is their requirement for training data that exemplifies all the information to be learned. This is a particular problem for visual question answering methods, which may be asked questions about virtually anything. The approach we propose is a step toward overcoming this limitation by searching for the information required at test time. The resulting method dynamically utilizes data from an external source, such as a large set of questions/answers or images/captions. Concretely, we learn a set of base weights for a simple VQA model, that are specifically adapted to a given question with the information specifically retrieved for this question. The adaptation process leverages recent advances in gradient-based meta learning and contributions for efficient retrieval and cross-domain adaptation. We surpass the state-of-the-art on the VQA-CP v2 benchmark and demonstrate our approach to be intrinsically more robust to out-of-distribution test data. We demonstrate the use of external non-VQA data using the MS COCO captioning dataset to support the answering process. This approach opens a new avenue for open-domain VQA systems that interface with diverse sources of data.
\end{abstract}

\section{Introduction}
\label{sec:intro}

\begin{figure}[t]
  \centering
  \includegraphics[width=0.87\linewidth]{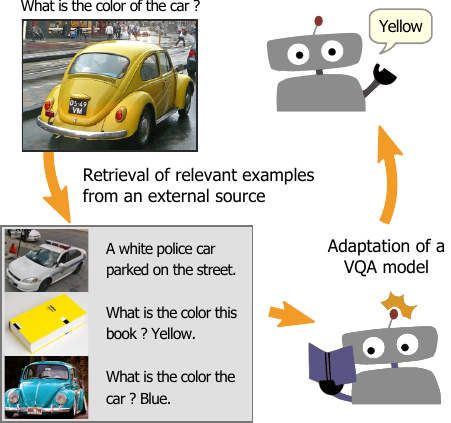}
  \label{fig:teaser}
  \vvspace{7pt}
  \caption{We propose a visual question answering (VQA) system able to retrieve and utilize information from an external source, at test time. The method learns to exploit external information of various forms, and we demonstrate question/answer tuples, but also images and corresponding captions. The method identifies the external information needed to answer a question and adapt its behaviour accordingly. This overcomes limitations of traditional approaches, including overfitting to the training data.}
\end{figure}

One of the ongoing criticisms of modern machine learning methods is that they presume the availability of large volumes of training data~\cite{marcus2018deep,yuille2018deep}. This training data should be representative of the distribution from which the test data will be sampled from, which may be unknowable at training time. These methods usually need constant retraining to accommodate recent data, or to alleviate under-generalizing under a domain shift between the training and test distributions.  While there exists a host of approaches to address these limitations (from continuum learning~\cite{thrun2012learning,thrun1998lifelong} to domain adaptation~\cite{chao2018crossvqa,saenko2010adapting,yosinski2014transferable} for example), the information extracted from the training data is typically fixed into the parameters of a model during training, and applied without modification thereafter. The approach we propose here addresses this limitation by exploiting new information as it comes to light, by seeking out relevant data from a large external data source. It actively adapts its behaviour according to the information gained from this data, which represents a fundamental change from pure supervised learning.

This paper demonstrates this novel capability on the task of Visual Question Answering (VQA). The task requires answering a previously unseen question about a previously unseen image. Questions are general and open-ended, and thus require a virtually unlimited array of information and skills to answer. The current approach to VQA is to train a neural network with end-to-end supervision of questions/answers (QAs). The supervised paradigm has been transformative for most classical tasks of computer vision, but it shows its limits on complex tasks that require more than pixel-processing and pattern recognition alone. VQA models trained in this fashion have revealed to rely mostly on biases and superficial correlations in the training data. For example, questions starting with ``How many...'' are usually answered with 2 or 3, and those starting with ``What sport ...'' with the answer tennis, which suffices to obtain high performance on benchmark datasets, where the training and test data are drawn from identical distributions.

The approach proposed in this paper is a step toward \emph{robust} VQA models, \ie capable of reasoning over visual and textual inputs, rather than regurgitating biases learned from a fixed training set. A robust evaluation of these capabilities has recently been made possible. Agrawal \etal~ proposed the VQA-CP (``changing priors'') dataset~\cite{vqacp}. In this resampled version of the VQA v2 dataset~\cite{goyal2016balanced}, the training and test sets are drawn from different distributions such that the question type (\ie the first few words such as ``What sport ...'' or ``How many ...'') cannot be relied upon to blindly guess the answer. The performance of existing methods significantly degrades in these conditions.


Our approach borrows ideas from recent research on meta learning~\cite{finn2017model,languageQueryMeta2018,teney2017meta}. So far, the ubiquitous approach to VQA has attempted to ``fit the world'' in a neural network, \ie capturing all of the information the method could ever require to answer any question within its weights. In contrast, we train a model to identify and utilize the relevant information from a external source of \emph{support data}. In the simplest instantiation of this principle, the support data is the training set of questions/answers itself, with the major novelty that it does not need to be fixed once the model is trained. The support data can expand at test time and could include data retrieved dynamically from live databases or web searches. The method then adapt itself dynamically using this data. To demonstrate the ability of the model to utilize non-VQA data (\ie other than QA tuples), we use the MS COCO captioning dataset~\cite{lin2014microsoft,chen2015microsoft} as a source of support data. While VQA data is expensive to acquire, captioned images are omnipresent on the web, and the ability to leverage such data is itself a major contribution.

The evaluation of our approach on VQA-CP demonstrates advantages over classical methods. It generalizes better and obtains state-of-the-art performance on the out-of-distribution test data of VQA-CP. Moreover, the model, once trained on a given distribution of QAs, can successfully adapt to a different distribution of an alternate support set. This is demonstrated with a novel leave-one-out evaluation with VQA-CP. Our experiments clearly demonstrate that the model makes use of the support data at test time, rather than merely capturing biases and priors of a training set. Consequently, a model trained with our approach could, for example, be reused in another domain-specific application by providing it with a domain-specific support set. This possibility opens the door to systems that reason over vision and language beyond the limited domain covered by any given training set. 

\noindent
The contributions of this paper are summarized as follows.
\setlist{nolistsep,leftmargin=*}
\begin{enumerate}[noitemsep]
  \item We propose a new approach to VQA in which the model is trained to retrieve and utilize information from an external source, to support its reasoning and answering process. We consider three instantiations of this approach, where the support data is the VQA training set itself (as an evaluation comparable to traditional models), VQA data from a different distribution, and non-VQA image captioning data.

  \item We propose an implementation of this approach based on a simple neural network and a gradient-based adaptation, which modifies its weights using selected support data. The method is based on the MAML algorithm~\cite{finn2017model} with novel contributions for efficient retrieval and cross-domain adaptation.

  \item We evaluate the components of our model on the VQA-CP v2 dataset. We demonstrate state-of-the-art performance, benefits in generalization, and the ability to leverage varied sources of support data. The novelty of the approach over existing practices opens the door to multiple opportunities to future research on VQA and vision/language reasoning.
\end{enumerate}

\section{Related work}
\label{sec:related}

\paragraph{Visual question answering}
VQA has gathered significant interest in the past few years~\cite{antol2015vqa,wu2017survey} alongside other tasks combining vision and language such as image captioning~\cite{chen2015microsoft} or visual dialog~\cite{das2017dialog}, for example. The appeal of VQA to the computer vision community is to constitute a practical evaluation of deep visual understanding. Open-domain VQA requires the visual parsing of an image, the comprehension of a text question, and reasoning over multiple pieces of information from these two modalities. See~\cite{wu2017survey} for a survey of modern methods and available datasets.

The ubiquitous approach to VQA is based on supervised learning. It is framed as a classification task over a large set of possible answers, and a machine learning model is optimized over a training set of human-provided questions and answers~\cite{antol2015vqa,goyal2016balanced,krishnavisualgenome,zhu2015visual7w}. Beyond apparent success on VQA benchmarks~\cite{fukui2016multimodal,teney2017challenge}, the approach was revealed to have severe limitations. The models following this approach prove to be overly reliant on superficial statistical regularities in the training sets, and their performance drops dramatically when evaluated on questions drawn from a different distribution~\cite{vqacp}, or on questions containing words and concepts that appear infrequently in the training data~\cite{rama2017vqa,teney2016zsvqa}. Popular benchmarks for VQA~\cite{antol2015vqa,goyal2016balanced} have involuntarily encouraged the development of methods that learn and leverage statistical patterns such as biases (\ie the long-tailed distributions of answers) and question-conditioned biases (which make answers easy to guess given a question without the image). These models can essentially bypass the steps of reasoning and image understanding that initially motivated research on VQA.

\paragraph{Robust evaluation of VQA}
Improved evaluations settings have recently been proposed. In~\cite{agrawal2017c,goyal2016balanced,zhang2015balanced}, the authors introduced balanced pairs of questions, \ie associating each question with a pair of images that lead to different answers. This procedure, however, had limited benefits. The usual metric of accuracy over individual questions still encouraged to learn and rely on the non-uniform distribution of answers, and the crowd-sourcing procedure used to gather balanced pairs introduced many irrelevant and nonsensical questions to the dataset.

Other recent proposals follow the idea of drawing the training and evaluation questions from different distributions. This discourages overfitting to statistical regularities specific to the training set. In~\cite{rama2017vqa,teney2016zsvqa}, the authors evaluate questions containing words and concepts that appear rarely in the training data. In~\cite{vqacp}, Agrawal \etal propose the VQA-CP dataset (for ``changing priors''), in which they enforce different training/test distributions of answers conditioned on the first few words of the question (\eg ``What is the color ...'' or ``How many ...''). Our experiments are conducted on VQA-CP as it represents the most challenging setting currently available.

\paragraph{Robust models for VQA}
The above robust evaluations have essentially pointed at the inadequacy of current approaches~\cite{vqacp,goyal2016balanced,rama2017vqa,teney2016zsvqa,zhou2015simple}. To address some of these these shortcomings, Agrawal \etal~\cite{vqacp} proposed a modular architecture that prevent it from relying on undesirable biases and priors in the training data. Ramakrishnan \etal~\cite{ramakrishnan2018overcoming}, introduced an information-theoretic regularizer to encourage the model to utilize the image by outperforming a ``blind'' guesser. In~\cite{teney2017meta}, Teney \etal proposed a meta learning approach to VQA that improved the recall on rare answers. Their work is the most relevant to this paper, although the methods differ significantly. We use a gradient-based adaptation procedure that update the weights of a whole VQA model, whereas \cite{teney2017meta} applied existing meta learning algorithms on the final classifier of a simple VQA model. We also formulate the use of support data as a retrieval task, whereas \cite{teney2017meta} processes the entire support set at every iteration, which is computationally challenging and the evaluation only include small-scale experiments. \cite{teney2017meta} is also to limited to QAs as support data, where our method is much more general.


\begin{figure*}[t]
  \centering
  \includegraphics[width=0.90\textwidth]{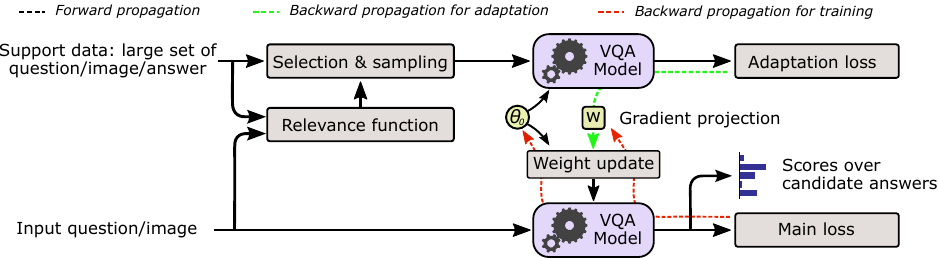}
  \vvspace{5pt}
  \caption{Data flow in the proposed method, using questions/answers as support data (\sect\ref{sec:qas}). The input question serves to retrieve pertinent instances from the support data using a relevance function. These instances are passed through the underlying VQA model (\fig\ref{fig:vqa}) to compute the adaptation loss $\loss_A$, using their ground truth answers. The gradient of the adaptation loss is backpropagated to the weights $\theta_0$ of the VQA model, which are updated, effectively adapting (\ie fine-tuning) the VQA model to the selected support examples. The input question is finally passed through the adapted model to predict scores for the final answer. During training, the gradient of the loss $\loss_M$ on the final predictions is backpropagated to optimize the pre-adaptation weights $\theta_0$ and the gradient projection (in yellow).}
  \label{fig:overall}
\end{figure*}

\paragraph{Meta learning}
Our central idea is to adapt a VQA model to each given question to incorporate additional information from an external source. The adaptation is implemented with the MAML meta learning algorithm~\cite{finn2017model}. Meta learning  or ``learning to learn'' \cite{naik1992meta,schmidhuber1987evolutionary,thrun2012learning} is a general paradigm to learn to build and/or update machine learning models, \eg to fine-tune the weights of a neural network~\cite{bengio1990learning,bengio1992optimization,schmidhuber1992learning}. Recent works in the area have focused on the adaptation of neural networks for few-shot image recognition~\cite{andrychowicz2016learning,finn2017model,hochreiter2001learning,ravi2016optimization}. MAML serves to identify a set of weights that can best serve as initial values, before adaptation through one or a few steps of gradient descent. In \cite{oneShotimitation2018a,oneShotimitation2018b}, the authors extended MAML to handle support data from a distinct domain, for robotic imitation learning from demonstration videos. We follow a similar idea to transform the gradients of a loss on captioning data into gradients suitable to update a VQA model. In \cite{languageQueryMeta2018}, Huang \etal turn the supervised task of language-to-query generation into a meta learning task. They introduce the concept of relevance functions to sample the training set. The approach is similar in spirit to our reformulation of VQA as a meta learning task. However, their aim is to improve accuracy by using specialized adapted models, while our objective is broader, as we also aim to leverage additional (non-VQA) sources of data.

\paragraph{Additional sources of data for VQA}
The limitations of the mainstream approach to VQA stem from the limited capacity of the training set and of the trained models. Instead of attempting to capture all the training information within the weights of a network, we use an external source of data that is not fixed after training. The capacity and capabilities of the model are thus essentially unbounded. Previous works~\cite{wu2015ask,wang2015explicit} have interfaced VQA models with knowledge bases, using ad hoc techniques to incorporate external knowledge. In comparison, this paper presents a more general approach, applicable to various types of support data. In~\cite{teney2016zsvqa,teney2017challenge}, the authors used web image search to retrieve visual representations of question and answer words. These representations are however optimized along the other weights of the network and fixed once trained. Recent works on text-based question answering used reinforcement learning to optimize the retrieval of external information~\cite{buck2017ask,narasimhan2016improving,nogueira2017task}, which is potentially complementary to our approach.


\section{Proposed approach}

Our central idea is to learn a VQA model that can subsequently adapt to each particular given question, using additional support data relevant to the question. Intuitively, the adaptation makes the VQA model specialized to the narrow domain of each question. The support data relevant to each question is retrieved dynamically from an external source (\fig\ref{fig:teaser}), which is assumed to be non-differentiable and/or too large to be processed all at once. Concretely, the support data can be the VQA training set itself (making evaluation comparable with traditional methods) but we also demonstrate the use of training QAs from a different distribution (Tables~\ref{tab:ablations2}--\ref{tab:vqacp2}), and the use of an image captioning dataset (\sect\ref{sec:exp}).

\subsection{Underlying VQA model}
Our approach builds around a standard VQA model that underlies the adaptation procedure. Formally, we denote with $\bx=\{\bq,\bv\}$ the input to the VQA model, made of the question $\bq$ (a string of tokens, each corresponding to a word) and of visual features $\bv$ pre-extracted from the given image (a feature map produced by a pre-trained convolutional neural network). The VQA model is represented as the function $f_\theta$ of parameters $\theta$. It maps $\bx$ to a vector of scores with $f_\theta(\bx)=\hat{\bs}$. The vector $\hat{\bs} \in [0,1]^A$ contains the scores predicted over $A$ candidate answers, typically the few thousands most frequent in the training set. The final answer is the one of largest score, $\argmax \hat{s}$. We denote with $\bs$ the vector of ground truth scores (which may contain multiple non-zero values when multiple answers are annotated as correct). 

\begin{figure}[t]
  \centering
  \includegraphics[width=0.99\linewidth]{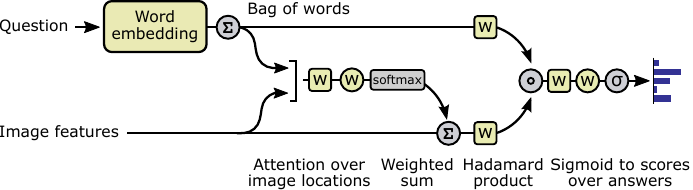}
  \vvspace{7pt}
  \caption{The simple VQA model underlying our method. It implements a classical joint embedding approach~\cite{teney2017challenge}. Yellow elements contain learnable weights. Circled and squared `{\texttt{w}}'s represent affine and non-linear projections, respectively. The above network is instantiated twice in the overall diagram of \fig\ref{fig:overall}.}
  \label{fig:vqa}
\end{figure}

The function $f$ is implemented as a neural network and $\theta$ denotes the set of all of its weights. Our contributions are not specific to any specific implementation of $f$. In practice, it corresponds to a classical joint embedding model~\cite{teney2017challenge} illustrated in \fig\ref{fig:vqa}. The network encodes the question as a bag-of-words, taking the average of learned word embeddings. It uses a single-headed, question-guided attention over image locations, a Hadamard product to combine the two modalities, and a non-linear projection followed by a sigmoid to obtain the scores $\hat{\bs}$. See Appendix~\ref{appendix:vqa} for details.

\subsection{Gradient-based adaptation}
\label{sec:qas}


The role of the adaptation procedure is to modify the weights of the VQA model to best tailor its capabilities to a given input question. The motivation for a specialized model is to be potentially be more effective than a general one for a same capacity of the underlying model. Our adaptation procedure is based on MAML~\cite{finn2017model}. The original MAML algorithm is designed for adaptation using support data of the same form as for task of interest, \ie questions with their ground truth answers. In \sect\ref{sec:captions}, we describe an extension to use support data of another task/domain.

The adaptation procedure takes in a set of support elements $\setSu\,$=$\,\{ \bx_j \}_j$ and base parameters $\theta_0$, which it adapts to $\theta_T$ over a small number $T$ of updates. The update rule is a gradient descent of step size $\alpha$:
\begin{align}
  \theta_{i+1} ~=~ \theta_i - \alpha \; \Sigma_j \nabla_{\theta_i} \loss_A(\bs_j, f_{\theta_i}(\bx_j))
  \label{eq:adaptation}
\end{align}
where $\loss_A$ is the adaptation loss which evaluates the predictions of the VQA model on the support data. In this case, $\loss_A$ is the binary cross-entropy loss typical used to train VQA models~\cite{teney2017challenge}. The above adaptation is performed when evaluating a given question at both training and test time. The key to benefit from this approach is to learn base parameters $\theta_0$ that are the most generally and most easily adaptable. They are optimized for the following objective:
\begin{align}
  \min_{\theta_0} \sum_{\bx_k \sim \setTr} \loss_M\big(\bs_k, f_{\theta_N}(\bx_k)\big)
  \label{eq:obj}
\end{align}
where the elements $(x_k,\bs_k)$ are drawn from a training set $\setTr$, and $\loss_M$ is the main loss on the VQA model (also called ``meta loss''~\cite{finn2017model}) that corresponds again to a binary cross-entropy. The objective can be optimized with standard backpropagation and stochastic gradient descent~\cite{finn2017model}. To avoid the expensive differentiation through the $T$ steps of adaptation (\eq\ref{eq:adaptation}), we use a first-order approximation of the gradient as in~\cite{nichol2018a}. The update rule is then
\begin{align}
  \theta_0 ~~\leftarrow~~ \theta_0 - \alpha' \; \nabla_{\theta_{T-1}} \loss_M(\bs_k, f_{\theta_T}(\bx_k)) ~.
  \label{eq:metaUpdate}
\end{align}
were $\alpha'$ is the learning rate. The whole procedure to evaluate any training or test instance is summarized as Algorithm~\ref{alg}. It is wort emphasizing that during training, a support set must be simulated to best mimic the conditions in which the model will be evaluated. If the support set is held constant during training, it would be treated as a static input, and the model is unlikely to generalize to different support data at test time. Therefore, it is crucial to present randomly sampled instances from the support set across the iterations in Algorithm~\ref{alg}.


\setlength{\textfloatsep}{0pt}

\begin{algorithm}[t]
  \KwIn{~~~Test or training instance $\bx\,$=$\,(\bq,\bv)$}
  ~~~~~~~	~~~~~~~~Support set ~$\setSu\,$=$\,\{ \bx_j \}_j$ ~with~~ $\bx_j$=$(\bq_j,\bv_j) ~\forall j$\\
  \KwOut{Vector of scores $\bs$ over candidate answers}
    //~\textit{Retrieve support relevant to $\bx$}:\\
    $\setSu_\bx \leftarrow \{\bx'_j\}_{j}^K \subset\setSu$ with max. precomputed $r(\bx,\bx'_j)$\\
    \For{$i$=0 to $(T-1)$}{\vspace{-10.5pt}~~~~~~~~~~~~~~~~~~~~~~~~~~~~~~~~~~//~\textit{For each adaptation step}\\
      $\setSu'_\bx \leftarrow ~K'$ random elements $\in \setSu_\bx$\\
      $\hat{\bs}'_j ~\leftarrow f_\theta(\bx'_j)$~~$\forall \bx_j \in \setSu'$~~//~\textit{Forward prop.}\\
      $\bd ~\leftarrow \Sigma_j \,\nabla_\theta \loss_A(\hat{\bs}'_j)$~~//~\textit{Backprop. adaptation loss}\\
      $\bd' ~\leftarrow g(\bd)$~~//~\textit{Gradient projection}\\
      $\theta_{i+1} ~\leftarrow \theta_i - \bd'$~~//~\textit{Update weights of VQA model}\\
    }
    $\hat{\bs} ~\leftarrow f_{\theta_T}(\bx)$~~//~\textit{Forward prop. with updated weights}\\
    \If{training}{
      $\bd ~\leftarrow \nabla_{\theta_0} \loss_M(\hat{\bs})$~~//~\textit{Backprop. main loss}\\
      $\theta_0 ~\leftarrow \theta_0 - \alpha \; \bd$~~//~\textit{Update base weights}\\
    }
  \caption{Evaluation of a tr. or test instance\label{alg}.}
\end{algorithm}
\setlength{\textfloatsep}{22pt}

\subsection{Using non-VQA data as support}
\label{sec:captions}

We now extend he method to use support data other than VQA instances (questions/answers). We apply it to the particular case of images/captions, although the approach is more generally applicable. The challenge is now to produce beneficial updates to the weights $\theta$ without access to a loss on the target VQA model. In practice, the format of captioning data (images with text) facilitates the implementation, as we can use a similar neural network as the VQA model $f_\theta$ to process them. We define a model $f'_\theta$ similar to $f_\theta$ up to the Hadamard product (\fig\ref{fig:vqa}). The final projection to answers scores is now meaningless for captions.

The adaptation procedure now proceeds as follows. The captions are passed through $f'$ and its output $\bh$ (the Hadamard product) is passed to the alternative adaptation loss $\loss_{A'}=\|\bh\|_2^2$. This squared L2 norm can be interpreted as measuring the compatibility of the embeddings of the caption and of the image. It encourages embedding spaces to align across support images and their captions. Importantly, this loss does not involve ground truth labels or answers, but it allows differentiation with respect to the weights $\theta$\footnote{Weights in $\theta$ corresponding to the final layers of $f_\theta$ and not present in $f'_\theta$ receive zero gradients when differentiating through~$f'_\theta$.}. The resulting gradients, however, cannot be assumed to be directly suitable to update the VQA model. We therefore pass them through a learned projection as $g(\nabla_\theta\loss_M)$. This produces gradients that can be plugged into \eq\ref{eq:adaptation} that now becomes
\begin{align}
  \theta_{i+1} &~=~ \theta_i - \alpha \; \Sigma_j~ g\big(\nabla_{\theta_i} \loss_{A'}(g_{\theta_i}(\bx_j))\big) ~.
  \label{eq:adaptation2}
\end{align}
The projection $g(\cdot)$ is implemented as a non-linear layer that is learned similarly to $\theta_0$, \ie by backpropagating the gradient of the main loss $\loss_M$ as in \eq\ref{eq:metaUpdate} (see details in the supplementary material).

\subsection{Retrieval of relevant support data}
\label{sec:retrieval}

The above descriptions assumed the availability of a set $\setSu_\bx$ of support examples relevant to an input question $\bx$. In our experiments, the support data $\setSu$ is the training split of a large VQA or captioning dataset. The selection of a relevant subset from $\setSu$ is a crucial step to make the model adaptation both efficient (by processing a much smaller subset $\setSu_bx$) and effective (by focusing the adapted model on a narrow domain around $\bx$). The method described below provides the adaptation algorithm with a subset of the support data of bounded size, and ensures its constant time complexity.

We formalize the retrieval process from $\setSu$ with a relevance function $r(\bx,\bx')$. It produces a scalar that reflects the pertinence of a support instance $\bx'=(\bq',\bv') \in \setSu$ to the input $\bx=(\bq,\bv)$. The top-$K$ elements $\{\bx'_j\}_j^K \subset \setSu$ of largest values $r(\bx,\bx')$ are identified, and then randomly subsampled to the set of $K'$ elements $\setSu_\bx=\{\bx'_j\}_j^{K'}$. 

The relevance function can in principled be learned using the gradient of the main loss $\nabla\loss_M$, although we did not explore this option. In our current implementation, we use a static relevance function that allows us to precompute its value between all training elements $\in \setTr$ and all elements of the simulated support set $\setSuTr$. This vastly improves the computational requirements during the training process. Our experiments evaluate conjunctions (products) of the following options:
\begin{equation}
\begin{aligned}
  r_0(\bx,\bx') &= 1~~~~\textrm{(Uniform sampling)}\\
  r_1(\bx,\bx') &= \textrm{number of common words between} ~\bq~ \textrm{and} ~\bq'\\
  r_2(\bx,\bx') &= 1 ~~\textrm{iff}~~ \bq' \textrm{contains word matching one of top-5}\\
                &~~~~~~~~~\textrm{answers from baseline VQA model on}~\bx.\\
                &= 0 ~~\textrm{otherwise}\\
  r_3(\bx,\bx') &= \big(\Sigma\bv / \|\Sigma\bv\|^2\big)~.~\big(\Sigma\bv' / \|\Sigma\bv'\|^2\big) ~~~\textrm{(Similarity}\\
                \textrm{of~} &\textrm{globally-pooled, L2-normalized image features)}.
  \label{eq:relevance}
\end{aligned}
\end{equation}
Note that the retrieval process could alternatively be formulated as a reinforcement learning task. This would allow optimizing the retrieval from ``black box'' data sources, such as web searches and dynamically-expanding databases~\cite{buck2017ask,narasimhan2016improving,nogueira2017task}, which we leave for future work.


\section{Experiments}
\label{sec:experiments}

We conducted extensive experiments to evaluate the contribution of the components of our method, and to compare its performance to existing approaches. We use the VQA-CP v2 dataset~\cite{vqacp}, which is the most challenging benchmark available. Its training and test splits have different distributions of answers conditioned on the first few words of the question, and was built by resampling the VQA v2 dataset~\cite{goyal2016balanced}. We hold out 8,000 questions from the VQA-CP training data to use as a validation set. All models are trained to convergence (with early stopping) on this validation set. Our underlying VQA model is a reimplementation of \cite{teney2017challenge} (see supplementary material for details). Experiments using captions as support data use the COCO captioning dataset~\cite{lin2014microsoft}. Since VQA-CP is itself made of images from COCO, we ensure that the captioned images also present in the VQA-CP test set are never used as support (neither during training nor evaluation). Please consult the supplementary material for additional implementation details and results. All results are reported using the standard VQA accuracy metric and broken down into the categories `yes/no', `number', and `other' as in~\cite{goyal2016balanced}.


\subsection{Results}
\label{sec:exp}


\paragraph{Contribution of the proposed components}
We first evaluate the impact of the proposed components with an ablative study (\tab\ref{tab:ablations}). For readability and computational reasons we focus on `other'-type questions\footnote{We focus on `other'-type questions because random guessing on the yes/no/number questions (or a buggy implementation !) does better than the best model in~\cite{vqacp}. We measured that random guessing achieves 72.9\% on yes/no questions (\cite{vqacp} gets 65.5\%) and random guessing of one/two achieves 34.1\% on `number' questions (\cite{vqacp} gets 15.5\%). This makes them unreliable for a meaningful analysis.} with a slightly simplified VQA model. Implementation details are provided in the supplementary material. We examine in \tab\ref{tab:ablations} a series of progressively more elaborate models. Each row corresponds to two different trained models, one trained for QAs as support (evaluated in the first 3 columns), another for captions (evaluated in the last column). All models using adaptation significantly outperform the baseline (first row). Interestingly, the optimal relevance function vary across the models for QAs and captions. The relevance function that includes the image similarity is only moderately useful, while the number of words in common between the question and the support text (QA or caption) proves very effective. Interestingly, in the case of captions, a uniform sampling already gives a clear improvement over the baseline model, but not with QAs, which we explain by the smaller size of the support set of captions.


We report results on both our validation set (of similar distribution as training data) and on the official test set (of different distribution). The overall lower performance on the latter shows the challenge of dealing with out-of-distribution data. The improvement in performance is much clearer on the test set than on the validation set. This demonstrates our contribution to improving generalization --~arguably the most challenging aspect of VQA~-- which is a significant side-effect of our adaptation-based approach.

\paragraph{Using image captions as support data}
We trained separate models for adaptation to questions/answers and to captions (\tab\ref{tab:ablations} last column). While performance improves over the baseline in both cases, the adaptation using QAs provides a bigger boost, given their direct relevance to the VQA task. The improvement by adaptation to captions demonstrates the ability of the method for picking up relevant information from non-VQA data, which opens a significant avenue for future work. This evaluation currently considers either QAs or captions separately. The combination of the two implies a number of non-trivial design decisions that we will explore in future work.

\begin{figure}[t]
  \centering
  \includegraphics[width=0.9\linewidth]{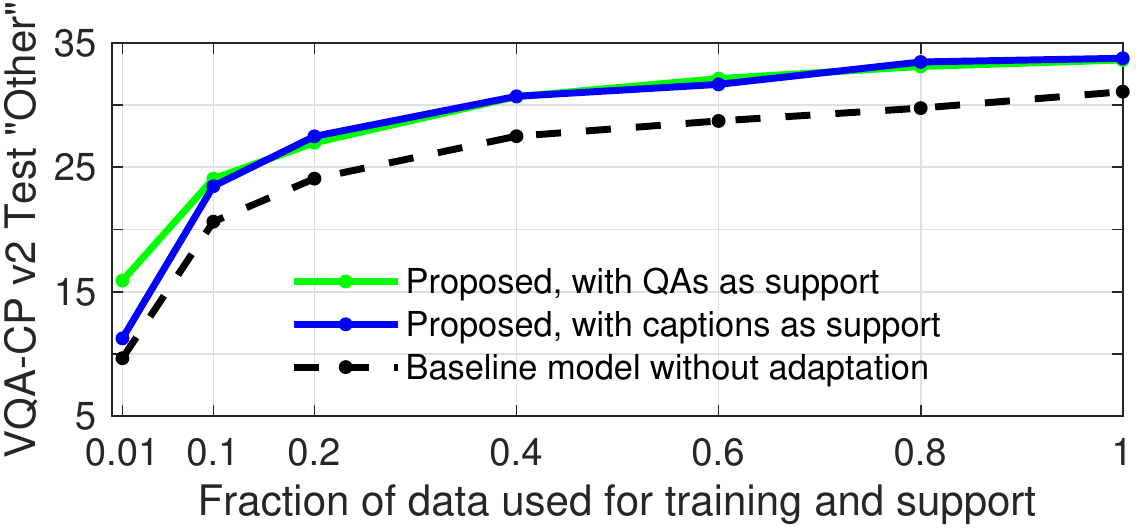}
  \vvspace{3pt}
  \caption{Accuracy as a function of the amount of data used.}
  \label{fig:nbSupport}
  \vspace{-10pt}
\end{figure}

\paragraph{Amount of retrieved support data}
In \fig\ref{fig:nbSupport}, we examine the performance of the model as a function of the amount of data it is trained with. To make the analysis comparable to the baseline VQA model, the support QAs are the same set of QAs as used for the training (of the baseline and of our model). In the case of captions, we use the same QAs for training, and a similarly subsampled set of captions as support data. We observe that our model is clearly superior to the baseline in all regimes, using both QAs or captions. The gain in performance is maintained even when the model is trained with very little data, in particular when using adaptation with QAs (using as little as 1\% of the whole training set).

Unfortunately,	the gains in using captions as support data levels off as the amount of support data increases (\fig\ref{fig:nbSupport}) and the performance does not surpass that obtained with QAs. One would rather hope continuing improvement as the model is provided with increasing amounts of support data. We believe that our current results do not prevent this prospect, and that the saturation stems from the particular distribution of captions in COCO. These captions are purely visual and descriptive, and they only cover a limited variety of concepts. In contrast, visual questions often require common sense and knowledge beyond visual descriptions (e.g. Why is the guy wearing such a weird outfit ? Is this a healthy breakfast ?). Other sources of data, including free-form captions and paired image-text data from the web may be more suitable for this purpose.

\begin{table}[t]
\small
\renewcommand{\tabcolsep}{0.02em}
\renewcommand{\arraystretch}{1.15}
\centering
\newcolumntype{w}{>{}c}
\newcolumntype{g}{>{}c}
\begin{tabularx}{\linewidth}{Xc c cg}
\Xhline{1\arrayrulewidth}
\multicolumn{5}{r}{Accuracy on VQA-CP v2 ``Other''} \\ 
& Val. & ~ & \multicolumn{2}{c}{Test} \\
\noalign{\vskip1pt}\Xhline{1\arrayrulewidth}\noalign{\vskip1pt}
\rowcolor{white} Ours without adaptation & 45.46 & ~ & \multicolumn{2}{c}{31.09}\\
\noalign{\vskip1pt}\Xhline{1\arrayrulewidth}\noalign{\vskip1pt}
\rowcolor{white} Ours with adaptation &   QAs & ~ & ~~~~~~QAs~~~~~~ & Capt. \vspace{-2pt}\\
\rowcolor{white} and, as support data: &   Tr. & ~ & Tr. & COCO \\ 
\noalign{\vskip1pt}\Xhline{1\arrayrulewidth}\noalign{\vskip1pt}
\noalign{\vskip.7pt}
Uniform sampling $r$=$r_0$ & 46.15 & ~ & 31.33 & 34.00\\
Relevance function $r$=$r_1$ & 44.41 & ~ & 31.79 & 29.18\\
Relevance function $r$=$r_2$ & 46.49 & ~ & 31.76 & 33.73\\
Relevance function $r$=$r_3$ & 46.32 & ~ & 31.68 & 33.51\\
Relevance function $r$=$r_2r_3$ & 46.17 & ~ & 31.09 & \textbf{34.26}\\
Relevance function $r$=$r_1r_2r_3$ & \textbf{46.79} & ~ & \textbf{34.25} & 33.44\\
\Xhline{1\arrayrulewidth}
\end{tabularx}
\normalsize
\normalsize
\caption{Ablative evaluation of the proposed method (see discussion in \sect\ref{sec:exp}). Each row corresponds to two different models, trained respectively for QAs (columns 1--3) and for captions (column 4) as support data. Gray cells use additional data during evaluation (QAs from VQA-CP test set in a leave-one-out setting) or during training+evaluation (COCO captions).}
\vspace{-5pt}
\label{tab:ablations}
\end{table}

\paragraph{Comparison to existing methods}
\tab\ref{tab:vqacp} presents a comparison of our results with existing approaches. We obtain state-of-the-art performance by a large margin over existing models and over our baseline model without adaptation. However, using captions as support data and trained on all question types (\textit{number}, \textit{yes/no}, and \textit{other}), the model performs poorly. We hypothesized that evidence for the \textit{number} and \textit{yes/no} questions was difficult to extract from captions. We therefore trained a model with adaptation using only \textit{other} questions. This model performs significantly better and clearly improves over the baseline. We indeed observed that captions seldom include counts or numbers, which can explain why they do not help on the corresponding questions. In the case of binary questions, it is possible than a different relevance function could address the issue.



\begin{table}[t]
\footnotesize
\renewcommand{\tabcolsep}{0.24em}
\renewcommand{\arraystretch}{1.35}
\centering
\begin{tabularx}{\linewidth}{X cccc}
\Xhline{1\arrayrulewidth}
                                          & \multicolumn{4}{c}{VQA-CP v2 Test} \\ \cline{2-5}
                                          & Overall & Yes/no & Numbers & Other \\
\Xhline{1\arrayrulewidth}
SAN~\cite{yang2015stacked}  & 24.96 & 38.35 & 11.14 & 21.74\\
GVQA~\cite{vqacp} & 31.30 & 57.99 & 13.68 & 22.14\\
UpDown~\cite{teney2017challenge} & 39.06 & 62.41 & 15.12 & 34.47\\
UpDown + regularizer~\cite{ramakrishnan2018overcoming} & 42.04 & \textbf{65.49} & 15.87 & 36.60\\
\hline
Ours without adaptation & 40.71 & 52.22 & 11.85 & 42.88\\
\multicolumn{5}{l}{Ours with adaptation and, as support data:}\\ 
QAs~(VQA-CP tr.), {$r$=$r_1r_2r_3$} & \textbf{46.00} & 58.24 & \textbf{29.49} & \textbf{44.33}\\
Captions~(COCO), {$r$=$r_1r_3$} & 39.84 & 48.78 & 12.40 & 42.93\\
Captions, trained only on `Other' q. & -- & -- & -- & 43.95\\
\Xhline{1\arrayrulewidth}
\end{tabularx}
\normalsize
\normalsize
\caption{Comparison with existing methods (accuracy on \emph{VQA-CP v2}). Our method significantly improves over the comparable baseline (the same VQA model without adaptation) and obtains performance superior to all existing models. Gray cells are not directly comparable to others as they use additional data (as in \tab\ref{tab:ablations}).}
\vspace{-5pt}
\label{tab:vqacp}
\end{table}

\begin{figure*}[t]
\vspace{-4pt}
\scriptsize
\renewcommand{\tabcolsep}{0.5em}
\renewcommand{\arraystretch}{1.1}
\centering
\begin{tabularx}{\linewidth}{c ccc l}
\Xhline{1\arrayrulewidth}
Input question & \multicolumn{3}{c}{Random selection of retrieved support data} & Predicted scores\\
\hline\vspace{-8pt}\\
\vis{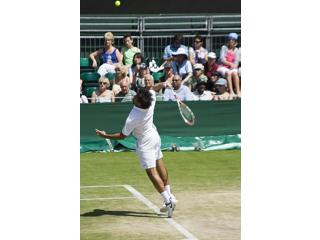}
\hline\vspace{-8pt}\\
\vis{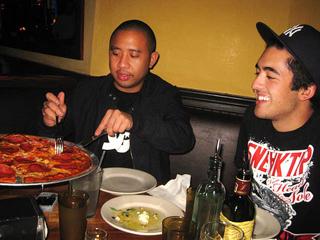}
\hline\vspace{-8pt}\\
\vis{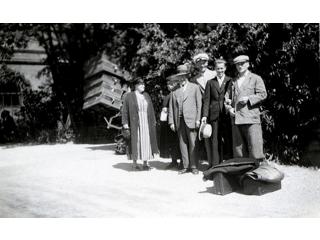}
\hline\vspace{-8pt}\\
\visCapt{71622}
\end{tabularx}

\normalsize
\vvspace{3pt}
\caption{Qualitative results comparing the top-5 answers and their scores predicted by the baseline, and by our model after adaptation. The retrieved support data (random samples are shown) is both visually and semantically relevant to each question.}

\label{fig:visualizations}
\vvspace{-8pt}
\end{figure*}

\paragraph{Qualitative results}
\fig\ref{fig:visualizations} presents results of our best models (using QAs or captions) with visualizations of support data sampled according to the relevance function. We observe that the retrieved support data is both semantically and visually relevant to each question.

Additional experiments and qualitative results are provided in the supplementary material.


\section{Conclusions}

We presented a new approach to VQA in which the model is trained to interface with an external source of data, and to use it to support its answering process. This is a significant departure from the classical training of a static model on a fixed dataset, which is obviously limited by finite capacity of the model and of the dataset. In contrast, our method retrieves information from the external source specifically for each given question. It then adapts the weights of its underlying VQA model, incorporating information from the external data, and specializing its capabilities to a narrow domain around the input question.

Our experiments demonstrate the benefits of the approach over existing models. It proves intrinsically more robust to out-of-distribution data, and it generalizes to different distributions when provided with novel support data. The model also introduces novel capabilities, in particular for leveraging non-VQA data (image captions) to support the answering process. This presents a number of opportunities to future research, for accessing ``black box'' data sources, such as web searches and dynamic databases. This opens the door to systems capable of reasoning over vision and language beyond the limited domain covered by any given training set.


\clearpage

{\small\bibliographystyle{ieee}\bibliography{Bibliography}}

\begin{thebibliography}{10}\itemsep=-1pt

\bibitem{vqacp}
A.~Agrawal, D.~Batra, D.~Parikh, and A.~Kembhavi.
\newblock Don't just assume; look and answer: Overcoming priors for visual
  question answering.
\newblock In {\em CVPR}, 2018.

\bibitem{agrawal2017c}
A.~Agrawal, A.~Kembhavi, D.~Batra, and D.~Parikh.
\newblock C-vqa: A compositional split of the visual question answering (vqa)
  v1. 0 dataset.
\newblock {\em arXiv preprint arXiv:1704.08243}, 2017.

\bibitem{anderson2017features}
P.~Anderson, X.~He, C.~Buehler, D.~Teney, M.~Johnson, S.~Gould, and L.~Zhang.
\newblock Bottom-up and top-down attention for image captioning and vqa.
\newblock {\em arXiv preprint arXiv:1707.07998}, 2017.

\bibitem{andrychowicz2016learning}
M.~Andrychowicz, M.~Denil, S.~Gomez, M.~W. Hoffman, D.~Pfau, T.~Schaul, and
  N.~de~Freitas.
\newblock Learning to learn by gradient descent by gradient descent.
\newblock In {\em Advances in Neural Information Processing Systems}, pages
  3981--3989, 2016.

\bibitem{antol2015vqa}
S.~Antol, A.~Agrawal, J.~Lu, M.~Mitchell, D.~Batra, C.~L. Zitnick, and
  D.~Parikh.
\newblock {{VQA}: Visual Question Answering}.
\newblock In {\em {Proc. IEEE Int. Conf. Comp. Vis.}}, 2015.

\bibitem{bengio1992optimization}
S.~Bengio, Y.~Bengio, J.~Cloutier, and J.~Gecsei.
\newblock On the optimization of a synaptic learning rule.
\newblock In {\em Preprints Conf. Optimality in Artificial and Biological
  Neural Networks}, pages 6--8. Univ. of Texas, 1992.

\bibitem{bengio1990learning}
Y.~Bengio, S.~Bengio, and J.~Cloutier.
\newblock {\em Learning a synaptic learning rule}.
\newblock Universit{\'e} de Montr{\'e}al, D{\'e}partement d'informatique et de
  recherche op{\'e}rationnelle, 1990.

\bibitem{buck2017ask}
C.~Buck, J.~Bulian, M.~Ciaramita, A.~Gesmundo, N.~Houlsby, W.~Gajewski, and
  W.~Wang.
\newblock Ask the right questions: Active question reformulation with
  reinforcement learning.
\newblock {\em arXiv preprint arXiv:1705.07830}, 2017.

\bibitem{chao2018crossvqa}
W.-L. Chao, H.~Hu, and F.~Sha.
\newblock Cross-dataset adaptation for visual question answering.
\newblock In {\em The IEEE Conference on Computer Vision and Pattern
  Recognition (CVPR)}, June 2018.

\bibitem{chen2015microsoft}
X.~Chen, H.~Fang, T.-Y. Lin, R.~Vedantam, S.~Gupta, P.~Dollar, and C.~L.
  Zitnick.
\newblock {Microsoft COCO captions: Data collection and evaluation server}.
\newblock {\em arXiv preprint arXiv:1504.00325}, 2015.

\bibitem{das2017dialog}
A.~Das, S.~Kottur, K.~Gupta, A.~Singh, D.~Yadav, J.~M. Moura, D.~Parikh, and
  D.~Batra.
\newblock {V}isual {D}ialog.
\newblock In {\em Proceedings of the IEEE Conference on Computer Vision and
  Pattern Recognition}, 2017.

\bibitem{finn2017model}
C.~Finn, P.~Abbeel, and S.~Levine.
\newblock Model-agnostic meta-learning for fast adaptation of deep networks.
\newblock {\em arXiv preprint arXiv:1703.03400}, 2017.

\bibitem{oneShotimitation2018a}
C.~Finn, T.~Yu, T.~Zhang, P.~Abbeel, and S.~Levine.
\newblock One-shot visual imitation learning via meta-learning.
\newblock In {\em Conference on Robot Learning (CoRL)}, pages 357--368, 2017.

\bibitem{fukui2016multimodal}
A.~Fukui, D.~H. Park, D.~Yang, A.~Rohrbach, T.~Darrell, and M.~Rohrbach.
\newblock Multimodal compact bilinear pooling for visual question answering and
  visual grounding.
\newblock {\em arXiv preprint arXiv:1606.01847}, 2016.

\bibitem{goyal2016balanced}
Y.~Goyal, T.~Khot, D.~Summers-Stay, D.~Batra, and D.~Parikh.
\newblock Making the {V} in {VQA} matter: Elevating the role of image
  understanding in {V}isual {Q}uestion {A}nswering.
\newblock {\em arXiv preprint arXiv:1612.00837}, 2016.

\bibitem{hochreiter2001learning}
S.~Hochreiter, A.~S. Younger, and P.~R. Conwell.
\newblock Learning to learn using gradient descent.
\newblock In {\em International Conference on Artificial Neural Networks},
  pages 87--94. Springer, 2001.

\bibitem{languageQueryMeta2018}
P.~Huang, C.~Wang, R.~Singh, W.~Yih, and X.~He.
\newblock Natural language to structured query generation via meta-learning.
\newblock In {\em {HLT-NAACL}}, pages 732--738, 2018.

\bibitem{krishnavisualgenome}
R.~Krishna, Y.~Zhu, O.~Groth, J.~Johnson, K.~Hata, J.~Kravitz, S.~Chen,
  Y.~Kalantidis, L.-J. Li, D.~A. Shamma, M.~Bernstein, and L.~Fei-Fei.
\newblock Visual genome: Connecting language and vision using crowdsourced
  dense image annotations.
\newblock {\em arXiv preprint arXiv:1602.07332}, 2016.

\bibitem{lin2014microsoft}
T.-Y. Lin, M.~Maire, S.~Belongie, J.~Hays, P.~Perona, D.~Ramanan,
  P.~Doll{\'a}r, and C.~L. Zitnick.
\newblock {Microsoft COCO: Common objects in context}.
\newblock In {\em {Proc. Eur. Conf. Comp. Vis.}}, 2014.

\bibitem{marcus2018deep}
G.~Marcus.
\newblock Deep learning: A critical appraisal.
\newblock {\em arXiv preprint arXiv:1801.00631}, 2018.

\bibitem{naik1992meta}
D.~K. Naik and R.~Mammone.
\newblock Meta-neural networks that learn by learning.
\newblock In {\em Neural Networks, 1992. IJCNN., International Joint Conference
  on}, volume~1, pages 437--442. IEEE, 1992.

\bibitem{narasimhan2016improving}
K.~Narasimhan, A.~Yala, and R.~Barzilay.
\newblock Improving information extraction by acquiring external evidence with
  reinforcement learning.
\newblock {\em arXiv preprint arXiv:1603.07954}, 2016.

\bibitem{nichol2018a}
A.~Nichol, J.~Achiam, and J.~Schulman.
\newblock On first-order meta-learning algorithms.
\newblock {\em arXiv preprint arXiv:1803.02999}, 2018.

\bibitem{nichol2018reptile}
A.~Nichol, J.~Achiam, and J.~Schulman.
\newblock On first-order meta-learning algorithms.
\newblock {\em arXiv preprint arXiv:1803.02999}, 2018.

\bibitem{nogueira2017task}
R.~Nogueira and K.~Cho.
\newblock Task-oriented query reformulation with reinforcement learning.
\newblock {\em arXiv preprint arXiv:1704.04572}, 2017.

\bibitem{pennington2014glove}
J.~Pennington, R.~Socher, and C.~Manning.
\newblock {Glove: Global Vectors for Word Representation}.
\newblock In {\em Conference on Empirical Methods in Natural Language
  Processing}, 2014.

\bibitem{ramakrishnan2018overcoming}
S.~Ramakrishnan, A.~Agrawal, and S.~Lee.
\newblock Overcoming language priors in visual question answering with
  adversarial regularization.
\newblock 2018.

\bibitem{rama2017vqa}
S.~K. Ramakrishnan, A.~Pal, G.~Sharma, and A.~Mittal.
\newblock An empirical evaluation of visual question answering for novel
  objects.
\newblock {\em arXiv preprint arXiv:1704.02516}, 2017.

\bibitem{ravi2016optimization}
S.~Ravi and H.~Larochelle.
\newblock Optimization as a model for few-shot learning.
\newblock 2017.

\bibitem{saenko2010adapting}
K.~Saenko, B.~Kulis, M.~Fritz, and T.~Darrell.
\newblock Adapting visual category models to new domains.
\newblock In {\em European conference on computer vision}, pages 213--226.
  Springer, 2010.

\bibitem{schmidhuber1987evolutionary}
J.~Schmidhuber.
\newblock {\em Evolutionary principles in self-referential learning, or on
  learning how to learn: the meta-meta-... hook}.
\newblock PhD thesis, Technische Universit{\"a}t M{\"u}nchen, 1987.

\bibitem{schmidhuber1992learning}
J.~Schmidhuber.
\newblock Learning to control fast-weight memories: An alternative to dynamic
  recurrent networks.
\newblock {\em Neural Computation}, 4(1):131--139, 1992.

\bibitem{teney2017challenge}
D.~Teney, P.~Anderson, X.~He, and A.~van~den Hengel.
\newblock Tips and tricks for visual question answering: Learnings from the
  2017 challenge.
\newblock 2018.

\bibitem{teney2016zsvqa}
D.~Teney and A.~van~den Hengel.
\newblock Zero-shot visual question answering.
\newblock 2016.

\bibitem{teney2017meta}
D.~Teney and A.~van~den Hengel.
\newblock Visual question answering as a meta learning task.
\newblock 2017.

\bibitem{thrun1998lifelong}
S.~Thrun.
\newblock Lifelong learning algorithms.
\newblock In {\em Learning to learn}, pages 181--209. Springer, 1998.

\bibitem{thrun2012learning}
S.~Thrun and L.~Pratt.
\newblock {\em Learning to learn}.
\newblock Springer Science \& Business Media, 2012.

\bibitem{wang2015explicit}
P.~Wang, Q.~Wu, C.~Shen, A.~{van~den~Hengel}, and A.~Dick.
\newblock Explicit knowledge-based reasoning for visual question answering.
\newblock {\em arXiv preprint arXiv:1511.02570}, 2015.

\bibitem{wu2017survey}
Q.~Wu, D.~Teney, P.~Wang, C.~Shen, A.~Dick, and A.~van~den Hengel.
\newblock Visual question answering: A survey of methods and datasets.
\newblock {\em Computer Vision and Image Understanding}, 2017.

\bibitem{wu2015ask}
Q.~Wu, P.~Wang, C.~Shen, A.~Dick, and A.~v.~d. Hengel.
\newblock {Ask Me Anything: Free-form Visual Question Answering Based on
  Knowledge from External Sources}.
\newblock In {\em {Proc. IEEE Conf. Comp. Vis. Patt. Recogn.}}, 2016.

\bibitem{yang2015stacked}
Z.~Yang, X.~He, J.~Gao, L.~Deng, and A.~Smola.
\newblock {Stacked Attention Networks for Image Question Answering}.
\newblock In {\em {Proc. IEEE Conf. Comp. Vis. Patt. Recogn.}}, 2016.

\bibitem{yosinski2014transferable}
J.~Yosinski, J.~Clune, Y.~Bengio, and H.~Lipson.
\newblock How transferable are features in deep neural networks?
\newblock In {\em Advances in neural information processing systems}, pages
  3320--3328, 2014.

\bibitem{oneShotimitation2018b}
T.~Yu, C.~Finn, A.~Xie, S.~Dasari, T.~Zhang, P.~Abbeel, and S.~Levine.
\newblock One-shot imitation from observing humans via domain-adaptive
  meta-learning.
\newblock 2018.

\bibitem{yuille2018deep}
A.~L. Yuille and C.~Liu.
\newblock Deep nets: What have they ever done for vision?
\newblock {\em arXiv preprint arXiv:1805.04025}, 2018.

\bibitem{zeiler2012adadelta}
M.~D. Zeiler.
\newblock {ADADELTA:} an adaptive learning rate method.
\newblock {\em arXiv preprint arXiv:1212.5701}, 2012.

\bibitem{zhang2015balanced}
P.~Zhang, Y.~Goyal, D.~Summers{-}Stay, D.~Batra, and D.~Parikh.
\newblock Yin and yang: Balancing and answering binary visual questions.
\newblock In {\em {Proc. IEEE Conf. Comp. Vis. Patt. Recogn.}}, 2016.

\bibitem{zhou2015simple}
B.~Zhou, Y.~Tian, S.~Sukhbaatar, A.~Szlam, and R.~Fergus.
\newblock Simple baseline for visual question answering.
\newblock {\em arXiv preprint arXiv:1512.02167}, 2015.

\bibitem{zhu2015visual7w}
Y.~Zhu, O.~Groth, M.~Bernstein, and L.~Fei-Fei.
\newblock {Visual7W: Grounded Question Answering in Images}.
\newblock In {\em {Proc. IEEE Conf. Comp. Vis. Patt. Recogn.}}, 2016.

\end{thebibliography}
\clearpage

\appendix
\section*{Supplementary material}
\vspace{12pt}

\section{Implementation of underlying VQA model}
\label{appendix:vqa}

The VQA model within our method follows the general description of Teney \etal~\cite{teney2017challenge} as illustrated in \fig\ref{fig:vqa} in the main paper. One exception is in the question encoding, where we replace their gated recurrent unit (GRU) with a bag of words, \ie a simple average of word embeddings. The first reason is computational, to avoid the relatively slow evaluation of the unrolled GRU. The second reason is that we encountered instabilities in the training of the adaptation method with the GRU. We suspect this to be due to our first-order approximation of the MAML algorithm.

Most implementation details follow \cite{teney2017challenge}. In particular, the non-linear operations in the network use gated hyperbolic tangent units. We use the ``bottom-up attention'' features~\cite{anderson2017features} of size 36$\times$2048, pre-extracted and provided by Anderson \etal\footnote{https://github.com/peteanderson80/bottom-up-attention} The word embeddings are initialized as GloVe vectors~\cite{pennington2014glove} of dimension 300, then optimized with the same learning rate as other weights of the network. All activations except the word embeddings and their average are of dimension 256. The answer candidates are those appearing at least 20 times in the VQA v2 training set, \ie a set of about 2000 answers. The output of the network is passed through a logistic function to produce scores in $[0,1]$. The final classifier is trained from a random initialization, rather than the visual and text embeddings of \cite{teney2017challenge}. In our ablative and in-depth experiments (\tab\ref{tab:ablations}, \fig\ref{fig:nbSupport}, and \fig\ref{fig:supportData}), we use a slightly simplified model, where the ``top-down'' attention map over the image is uniform. The image features of size 36$\times$2048 are thus averaged uniformly to a vector of size 1$\times$2048. This significantly reduces the cost of training and evaluating the model since these averages can be precomputed and fit in memory for the whole dataset. The relevance function $r_3$ (\sect\ref{sec:retrieval}) also uses these global image features.

\section{Implementation of adaptation algorithm}
\label{appendix:adaptation}
We use the AdaDelta algorithm~\cite{zeiler2012adadelta} to train the model's weights ($\theta_0$ and those of the gradient projection) with backpropagation from the loss $\loss_M$. Following this practice, we also found it beneficial to replace the gradient descent step of the adaptation (\eq\ref{eq:adaptation} and \ref{eq:adaptation2}) with the AdaDelta weight update (see details in \cite{zeiler2012adadelta}). This effectively determines the size of the gradient step $\alpha$ automatically based on a rolling average of the weights' and gradients' magnitudes. This makes the weight updates much more stable, and it eliminates the hyperparameter $\alpha$.

The gradient projection $g_{\psi}(\cdot)$ implemented as a simple linear scaling, with no biases, and no cross-talk across dimensions. For example, to adapt a linear layer that uses weights $W \in \mathbb{R}^{256\times256}$, the gradient $\nabla_{W} \loss_A$ is transformed with
\begin{align}
  g_{\psi_M}(\nabla_{W} \loss_A) = \psi_M ~\circ~ \nabla_{W} \loss_A
  \label{eq:proj}
\end{align}
where $\psi_M \in \mathbb{R}^{256\times256}$ represents the parameters of the projection and $\circ$ the Hadamard (element-wise) product.

The adaptation algorithm uses a number $T$=3 updates during training and evaluation. This value was selected in 1--5 by cross-validation.

The whole method is trained with mini-batches of size 128. The evaluation also uses mini-batches (of the same size) in a transductive manner, \ie sharing information across multiple test instances, as done in existing implementations of MAML~\cite{finn2017model,nichol2018reptile}. This means that the adaptation algorithm effectively uses support data retrieved for 128 questions at a time. The primary reason for mini-batches during evaluation is computational, but we did not observe improvements in accuracy with smaller batch sizes (down to processing one single instance at a time), whether for training and/or evaluation.

\section{Additional experiments}

\subsection{Varying the amount of support data}

\begin{figure}[t]
  \centering
  \includegraphics[width=0.94\linewidth]{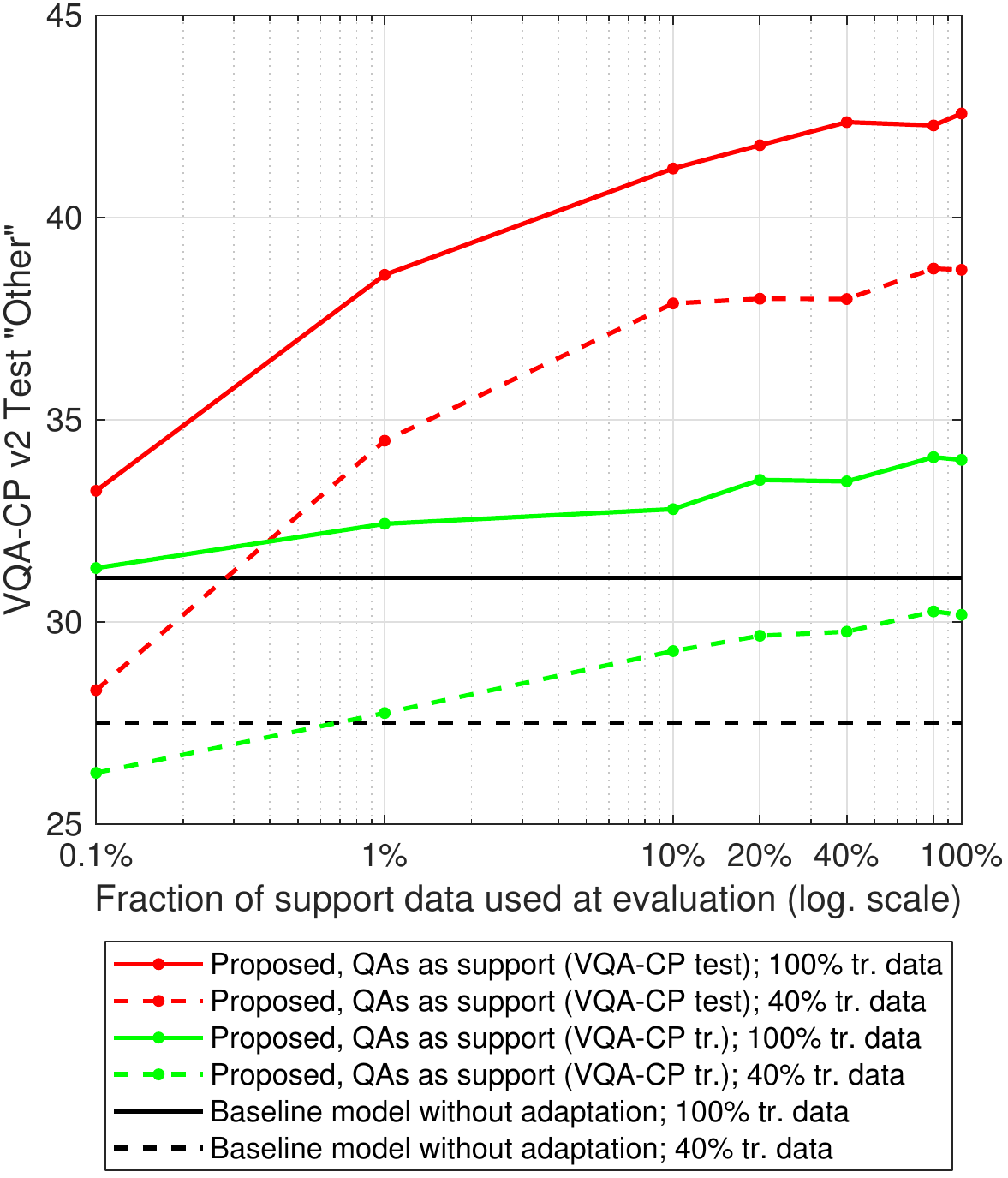}
  \vvspace{4pt}
  \caption{Varying the amount of support data used during evaluation.}
  \label{fig:supportData}
\end{figure}

We performed additional experiments in which in varied the amount of support data available during the evaluation of the model (\fig\ref{fig:supportData}). This serves to verify that the model makes actual use of information from the support data. We indeed observe that the performance increases as more data is made available. We repeated the experiment with a model initially trained with only 40\% of the data (dashed lines in \fig\ref{fig:supportData}). The trend of the accuracy versus the amount of support data remains similar. The overall performance is however lower. This indicates room for improvement for the adaptation algorithm. Ideally, a model trained with less data should approach the performance of a model trained with more data, when provided with this data (as support) at test time.


\subsection{Generalization to support from a different distribution}
We evaluated the proposed model by providing it with support data from a different distribution than the data it is originally trained with (Tables~\ref{tab:ablations2}--\ref{tab:vqacp2}). For these experiments, we use the VQA-CP in a ``leave-one-out'' setting: we use the test set itself as the support data, and masking the intersection of the support data with a test instance currently evaluated. More precisely, all QAs relating to the same image as the current test question are left out of the utilized support. The results of this experiment show that the model can very effectively adapt to this novel support data, as the accuracy gets a significant jump, approaching the performance of the validation set (which is of the same distribution as the initial training data). We suspected the increase in performance might be simply due to the larger amount of data (the original training data plus the additional test set provided as support). We disproved this hypothesis by repeating the experiment with a model trained with less initial training data and less support data, such as to match the same total amount of data provided to the baseline (details in the supplementary material). This experiment gave a similarly high accuracy, which demonstrates that the model is indeed capable of adapting on-the-fly to the provided support data, even when it significantly differs from the data it was originally trained with.

\section{Qualitative results}

We provide additional qualitative results in the following pages. A first set of results uses support data made of QAs. A second set uses support data made of captioned images (as indicated in column headings).

\begin{table}[h!]
\small
\renewcommand{\tabcolsep}{0.02em}
\renewcommand{\arraystretch}{1.2}
\centering
\newcolumntype{w}{>{}c}
\newcolumntype{g}{>{}c}
\begin{tabularx}{\linewidth}{Xcccc}
\Xhline{1\arrayrulewidth}
\multicolumn{5}{r}{VQA-CP v2 Test split, ``Other'' questions} \\ 
\noalign{\vskip1pt}\Xhline{1\arrayrulewidth}\noalign{\vskip1pt}
\rowcolor{white} Ours with adaptation & ~ &    ~~~~~QAs~~~~~ & ~ & ~~~~~QAs~~~~~\vspace{-2pt}\\
\rowcolor{white} and, as support data: & ~ &    Tr. & ~ & Test \\
\noalign{\vskip1pt}\Xhline{1\arrayrulewidth}\noalign{\vskip1pt}
Uniform sampling $r$=$r_0$ & ~ &  31.33 & ~ & 32.83\\
Relevance function $r$=$r_1$ & ~ &  31.79 & ~ & 37.19\\
Relevance function $r$=$r_2$ & ~ &  31.76 & ~ & 36.28\\
Relevance function $r$=$r_3$ & ~ &  31.68 & ~ & 33.52\\
Relevance function $r$=$r_2r_3$ & ~ &  31.09 & ~ & 37.78\\
Relevance function $r$=$r_1r_2r_3$ & ~ &  \textbf{34.25} & ~ & \textbf{43.52}\\
\Xhline{1\arrayrulewidth}
\end{tabularx}
\normalsize
\normalsize
\caption{Complement to Table~\ref{tab:ablations}. We evaluate the different versions of our model with, as support data, QAs from the training set (first column, identical to Table~\ref{tab:ablations}) and QAs from the test set (second column, in a leave-one-out protocol). These results are not comparable to competing models since they use more data, but the clear improvement in the second column demonstrates that the model clearly adapts to support data from a distribution different from the one it was trained with (since the support QAs now reflect the distribution of the test questions). We envision this capability to allow a pretrained VQA model to be applied to various domains by simply providing it, at test time, with domain-specific support data.}
\vspace{-8pt}
\label{tab:ablations2}
\end{table}
\begin{table}[h!]
\footnotesize
\renewcommand{\tabcolsep}{0.24em}
\renewcommand{\arraystretch}{1.2}
\centering
\begin{tabularx}{\linewidth}{X cccc}
\Xhline{1\arrayrulewidth}
                                          & \multicolumn{4}{c}{VQA-CP v2 Test split} \\ \cline{2-5}
                                          & Overall & Yes/no & Numbers & Other \\
\Xhline{1\arrayrulewidth}
\multicolumn{5}{l}{Ours with adaptation and, as support data:}\\ 
QAs~(VQA-CP tr.), {$r$=$r_1r_2r_3$} & {46.00} & 58.24 & {29.49} & {44.33}\\
QAs~(VQA-CP test), {$r$=$r_1r_2r_3$} & 52.09 & 62.02 & 47.66 & 48.21\\
\Xhline{1\arrayrulewidth}
\end{tabularx}
\normalsize
\normalsize
\caption{Complement to Table~\ref{tab:vqacp} (first row is identical to Table~\ref{tab:vqacp}). This demonstrates the same effect as explained for Table~\ref{tab:ablations2}.}
\vspace{-5pt}
\label{tab:vqacp2}
\end{table}
\begingroup
\let\clearpage\relax
\endgroup%
\onecolumn
\renewcommand{\tabcolsep}{0.5em}
\renewcommand{\arraystretch}{1.1}
\begin{tabularx}{\linewidth}{c ccc X}\hline Input question & \multicolumn{3}{c}{Samples of retrieved support data (QAs)} & Predicted scores\\\hline\vspace{-8pt}\\
\vis{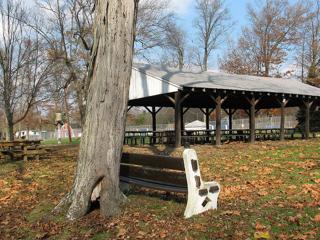}
\hline\vspace{-8pt}\\
\vis{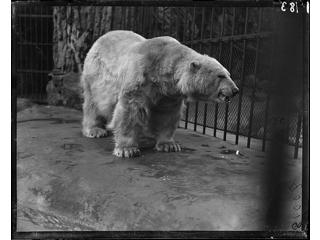}
\hline\vspace{-8pt}\\
\vis{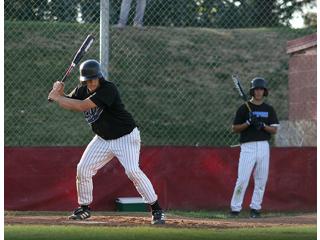}
\hline\vspace{-8pt}\\
\vis{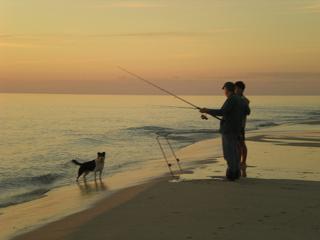}
\hline\vspace{-8pt}\\
\end{tabularx}
\newpage
\begin{tabularx}{\linewidth}{c ccc X}\hline\vspace{-8pt}\\
\vis{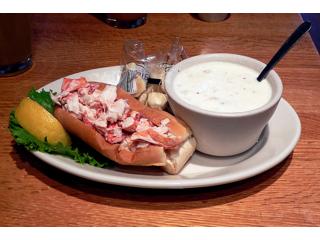}
\hline\vspace{-8pt}\\
\vis{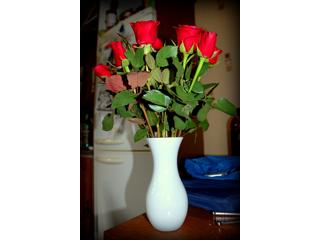}
\hline\vspace{-8pt}\\
\vis{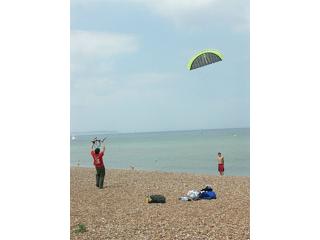}
\hline\vspace{-8pt}\\
\vis{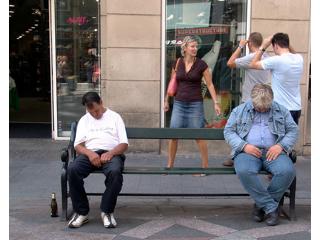}
\hline\vspace{-8pt}\\
\end{tabularx}
\newpage
\begin{tabularx}{\linewidth}{c ccc X}\hline\vspace{-8pt}\\
\vis{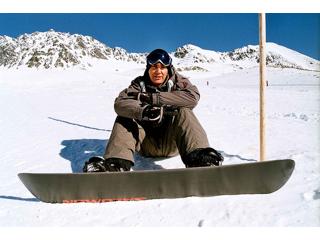}
\hline\vspace{-8pt}\\
\vis{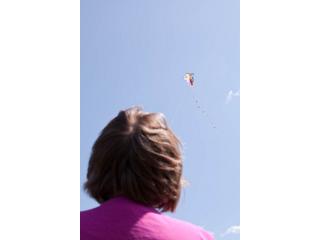}
\hline\vspace{-8pt}\\
\vis{59920}
\hline\vspace{-8pt}\\
\vis{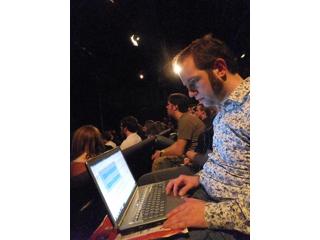}
\hline\vspace{-8pt}\\
\end{tabularx}

\renewcommand{\tabcolsep}{0.5em}
\renewcommand{\arraystretch}{1.1}
\begin{tabularx}{\linewidth}{c ccc X}\hline Input question & \multicolumn{3}{c}{Samples of retrieved support data (captions)} & Predicted scores\\\hline\vspace{-4pt}\\
\visCapt{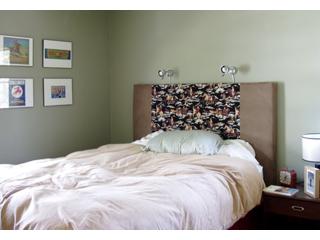}
\hline\vspace{2pt}\\
\visCapt{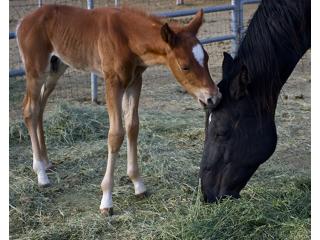}
\hline\vspace{2pt}\\
\visCapt{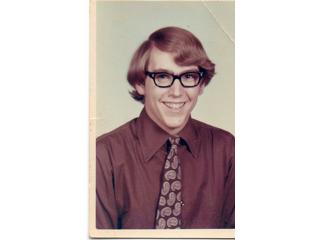}
\hline\vspace{2pt}\\
\end{tabularx}
\newpage
\begin{tabularx}{\linewidth}{c ccc X}\hline
\visCapt{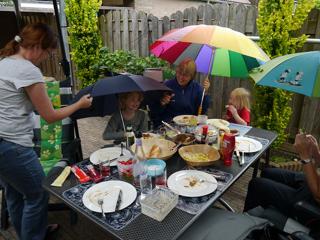}
\hline\vspace{-8pt}\\
\visCapt{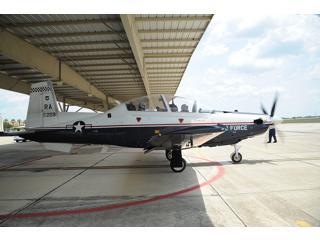}
\hline\vspace{-8pt}\\
\visCapt{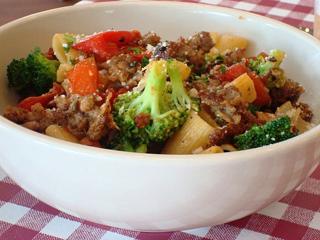}
\hline\vspace{-8pt}\\
\visCapt{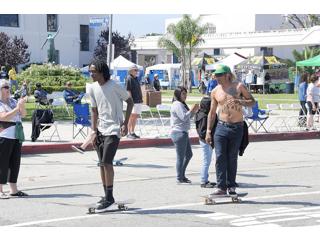}
\hline\vspace{-8pt}\\
\end{tabularx}
\newpage
\begin{tabularx}{\linewidth}{c ccc X}\hline
\visCapt{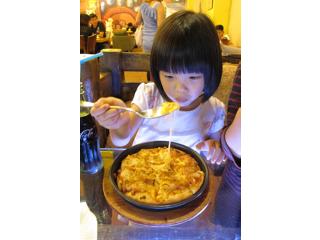}
\hline\vspace{-8pt}\\
\visCapt{388099}
\hline\vspace{-8pt}\\
\visCapt{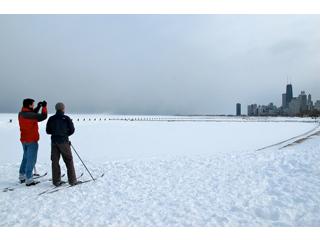}
\hline\vspace{-8pt}\\
\visCapt{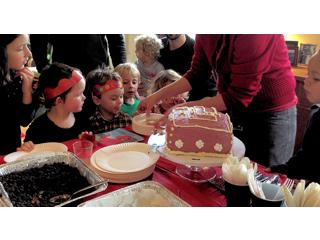}
\hline\vspace{-8pt}\\
\end{tabularx}%
\end{document}